\documentclass{article}
\usepackage{arxiv}
\usepackage[numbers,sort&compress]{natbib}
\usepackage[utf8]{inputenc} 
\usepackage[T1]{fontenc}    
\usepackage{ dsfont }
\usepackage{hyperref}       
\usepackage{url}
\usepackage{textcomp}
\usepackage{ gensymb }
\usepackage[ruled,vlined]{algorithm2e}
\usepackage{algorithmic}
\usepackage{booktabs}       
\usepackage{amsfonts}       
\usepackage{nicefrac}       
\usepackage{microtype}      
\usepackage{lipsum}
\usepackage{float}
\usepackage{graphicx}
\usepackage{multirow}
\usepackage{array}
\usepackage{comment}
\usepackage{longtable}
\usepackage{ragged2e}
\usepackage{subfig}
\usepackage{amsmath}  
\usepackage{amssymb}  
\title{BSGAN: a novel oversampling technique for imbalanced pattern recognitions}

\author{
  Md Manjurul Ahsan \\
  School of Industrial and Systems Engineering\\
  University of Oklahoma\\
  Norman, Oklahoma-73019 \\
  \texttt{ahsan@ou.edu} \\
   \And
 Shivakumar Raman \\
  School of Industrial and Systems Engineering\\
  University of Oklahoma\\
  Norman, Oklahoma-73019\\
  \texttt{raman@ou.edu}\\
  \And
 Zahed Siddique \\
  School of Aerospace and Mechanical Engineering\\
  University of Oklahoma\\
  Norman, Oklahoma-73019\\
  \texttt{zsiddique@ou.edu}} 

\begin{document}
\maketitle
\begin{abstract}
Class imbalanced problems (CIP) are one of the potential challenges in developing unbiased Machine Learning (ML) models for predictions. CIP occurs when data samples are not equally distributed between the two or multiple classes. Borderline-Synthetic Minority Oversampling Techniques (SMOTE) is one of the approaches that has been used to balance the imbalance data by oversampling the minor (limited) samples. One of the potential drawbacks of existing Borderline-SMOTE is that it focuses on the data samples that lied at the border point and gives more attention to the extreme observations, ultimately limiting the creation of more diverse data after oversampling, and that is the almost scenario for the most of the borderline-SMOTE based oversampling strategies. As an effect, marginalization occurs after oversampling. To address these issues, in this work, we propose a hybrid oversampling technique by combining the power of borderline SMOTE and Generative Adversarial Network to generate more diverse data that follow Gaussian distributions. We named it BSGAN and tested it on four highly imbalanced datasets— Ecoli, Wine quality, Yeast, and Abalone. Our preliminary computational results reveal that BSGAN outperformed existing borderline SMOTE and GAN-based oversampling techniques and created a more diverse dataset that follows normal distribution after oversampling effect.

\end{abstract}
\maketitle
\section{Introduction}
Imbalanced data classification is a problem in data mining domains where the proportion of data class of a dataset differs relatively by a substantial margin. In this situation, one class contains a few numbers of samples (known as the minor class), whereas the other class contains the majority of the samples~\citep{ahsan2022imbalanced, longadge2013class}. Such an imbalanced ratio produces biased results towards the minor class (minority classes). 
The issue of imbalanced data is a prevalent problem in many real-world scenarios, such as detecting fraudulent financial transactions, identifying rare medical conditions, or predicting equipment failures in manufacturing~\citep{sahu2020dual, jalali2019predicting}.
Several approaches have been introduced over the years, and among them, the most popular methods used for handling imbalanced data are neighborhood cleaning rule, cost-sensitive, and neural network algorithms. There are three major ways to handle Class Imbalanced Problems (CIP)~\citep{gosain2017handling, geng2019cost} :
\begin{itemize}
    \item Data level solutions (i.e., random undersampling, random oversampling, one-sided selection)
    \item Cost-sensitive (i.e., cost-sensitive resampling, cost-sensitive ensembles)
    \item Ensemble algorithms (i.e., boosting and bagging, random Forest)
\end{itemize}

\begin{figure} [h!]
    \centering
    \includegraphics[width=.5\textwidth]{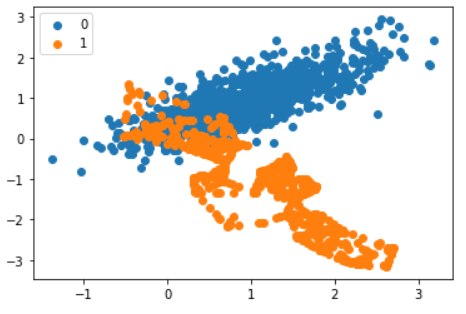}
    \caption{The Oversampling effect of SMOTE often creates noisy samples and, therefore, major and minor samples overlap. Here $0$ indicates the initial major samples and $1$ indicates minor samples after oversampling.}
    \label{fig:fig1}
\end{figure}

Among different data-level solutions, oversampling techniques are the most widely used, and the Synthetic Minority Oversampling Technique (SMOTE) is the most often adopted by researchers and practitioners to handle CIP. Chawla et al. (2002) initially proposed SMOTE-based solutions, and they became popular due to their capability to produce synthetic samples, ultimately leading to the opportunity to reduce the biases of the ML models~\citep{chawla2002smote}. However, the existing SMOTE has two potential drawbacks~\citep{ahsan2022imbalanced}:
\begin{enumerate}
    \item The synthetic instances generated by the SMOTE often are in the same direction. As an effect, for some of the ML classifiers, it is hard to create a decision boundary between the major and minor classes.
    \item SMOTE tends to create a large number of noisy data, which often overlaps with major class (as shown in Figure~\ref{fig:fig1}).
\end{enumerate}

To overcome the noise generated by the SMOTE, several expansion of SMOTE has been proposed, such as Support Vector Machine (SVM)-SMOTE, Safe-Level SMOTE, and Borderline-SMOTE. However, SVM-SMOTE is known for its sensitivity issues with multiclass data samples, while borderline SMOTE can only focus on the minor samples that are close between the boundaries and major class~\citep{han2005borderline}. 

Therefore, both SVM-SMOTE and Borderline-SMOTE have limitations in creating diverse and normally distributed data with less marginalization after data expansion. Considering these challenges, in this paper, we propose a hybrid method of oversampling that exploits the diverse sets of samples, which will be helpful for the ML-based model to differentiate between major and minor classes. Our hybrid approach combines two popular oversampling techniques: Borderline-SMOTE and GAN. First, we propose combining two CNN architectures—generator and discriminator—with Borderline-SMOTE into a single architecture that is trained end to end. Second, we provide the final prediction by averaging all the predictions. Our proposed approach is tested on four highly imbalanced benchmark datasets. 

Our main contributions can be summarized as follows:
\begin{itemize}
    \item We modified and designed the generator and discriminator networks and proposed an improved GAN model that can train with small data set to a large dataset for the binary classification.
    \item Later, we propose a new oversampling technique by combining Borderline-SMOTE and GAN, namely BSGAN.
    \item We propose a Neural Network (NN) model, which is later used to train and test datasets with and without oversampling.
    \item We implement and test the performance of Borderline-SMOTE, GAN, and BSGAN on four highly imbalanced datasets—Ecoli, Yeast, Winequality, and Abalone. Later, the performance of those three algorithms is compared with the dataset without oversampling in terms of accuracy, precision, recall, and F1-score.
    \item Finally, We compare our proposed BSGAN model performance with some of the reference literature. The preliminary findings revealed that our proposed approach outperformed many of the existing GAN-based oversampling approaches and can handle sensitive data issues. Our proposed model also creates a more diverse dataset that incorporates Gaussian distributions instead of creating extreme outliers as often produced by many existing methods.
\end{itemize}

The motivation of this study is to further improve the performance of data oversampling techniques by proposing a new approach that combines the advantages of Borderline-SMOTE and GAN. By exploring new ways to balance imbalanced datasets, this study seeks to provide valuable insights into improving the accuracy and effectiveness of ML models in a range of fields where imbalanced data is a common challenge.

The rest of the paper is organized as follows: Section~\ref{rw} covers some previously published research that focused on different approaches to handling CIP. In Section~\ref{met}, we provide a brief description of SMOTE, Borderline-SMOTE, GAN, and the architecture of the proposed BSGAN technique. In Section~\ref{res} performance of the various oversampling techniques is evaluated by considering various statistical measurements. An overall discussion and comparison with the current work have been summarized in Section~\ref{dis}, wherein Section~\ref{con} concludes the paper's contributions with potential remarks.

\section{Related work}\label{rw}
CIPs are one of the existing and ongoing research in data science domains. As the imbalanced ratio potentially affects the models' prediction, several approaches have been proposed to balance the dataset in a way that can be used to develop an unbiased prediction model~\citep{lango2022makes,fern2022multi}. Among them, oversampling approaches are most widely used as they provide data-level solutions with less complexity and computational issues~\citep{goodman2022distance}. Therefore, we have focused mainly on popular oversampling methods such as SMOTE, Borderline-SMOTE, and SVM-SMOTE and their modified, adopted versions that have been proposed during the last few years.

The marginalization and noise sensitivity issue of existing SMOTE and borderline-SMOTE has been addressed by many of the recent literature. For instance, Li et al. (2022) introduced cluster-Borderline-SMOTE, a hybrid method to classify rock groutability~\citep{li2022hybrid}. Ning et al. (2021) combined SMOTE with Tomek-links techniques for identifying glutarylation sites~\citep{ning2021novel}. Zhang et al. (2020) proposed a modified borderline-SOMOTE by combining it with the ReliefF algorithms for intrusion detection~\citep{zhang2020network}. Sun et al. (2020) applied ensemble techniques by combining Adaboost-SVM with SMOTE. The empirical experiments are carried out based on the financial data of 2628 Chinese listed companies~\citep{sun2020class}. Liang et al. (2020) introduced hybrid oversampling techniques by combining k-means and SVM. The authors claim that the proposed models can generate samples without considering the outlier samples~\citep{liang2020lr}. However, none of the experiments justifies how their proposed model creates a normally distributed dataset.

Recently, GANs have demonstrated the potential to create real samples using random noise~\citep{ahsan2022imbalanced}. For instance, the existing GAN can be utilized to create real images of any objects from random noise with several neural network iterations. While GAN is generally extensively applied in computer vision domains, the adoption of GAN can be observed in handling class imbalanced problems. For instance, Gombe et al. (2019) proposed Multi-scale Feature Cascade (MFC)-GAN, where multiple fake samples are used to create synthetic data to develop a balanced dataset~\citep{ali2019mfc}. Kim et al. (2020) used GAN-based approaches to detect anomalies from publicly available datasets like MNIST and Fashion MNIST~\citep{kim2020gan}. 

Rajabi et al. (2022) present a novel approach for generating synthetic data that balances the trade-off between accuracy and fairness through their proposed method, TabFairGAN. Their approach specifically focuses on complex tabular data and has been empirically evaluated on various benchmark datasets, including UCI Adult, Bank Marketing, COMPAS, Law School, and the DTC dataset. The results of the experiments reveal that TabFairGAN demonstrates promising performance, achieving an average accuracy of $78.3 \pm 0.001\%$ and an F1-score of $0.544 \pm 0.002$~\citep{rajabi2022tabfairgan}.
Engelmann and Lessmann (2021) proposed the cWGAN approach for generating tabular datasets containing both numerical and categorical data. The effectiveness of this approach was evaluated on several highly imbalanced benchmark datasets, including the German credit card, HomeEquity, Kaggle, P2P, PAKDD, Taiwan, and Thomas datasets. The results showed that the cWGAN approach achieved an overall rank of 4.1 for Logistic Regression~\citep{engelmann2021conditional}.
Jo and Kim (2022) presented the Outlier-robust (OBGAN) method for generating data from the minority region close to the border. The performance of the OBGAN method was evaluated on several UCI imbalanced datasets. The results indicated that the OBGAN method achieved the highest recall and F1-score of 0.54 and 0.65, respectively~\citep{jo2022obgan}.

However, most of the existing GAN-based approaches are computationally expensive and often hard to train due to their instability.

Considering this opportunity into account, in this work, we propose a novel hybrid approach by combing borderline SMOTE and GAN and named it BSGAN. The BSGAN is tested along with borderline SMOTE, GAN, and without oversampling on four highly imbalanced datasets— Ecoli, Wine quality, Yeast, and Abalone. The empirical, experimental results demonstrate that BSGAN outperformed most of the existing tested techniques regarding various statistical measures on most of the datasets used in this study.

Table~\ref{tab:compare} summarizes the literature that used GAN-based approaches to handle class imbalanced problems. It provides information on each study's contributions, algorithms, datasets, performance, misclassification evaluation, and algorithm complexity.

\begin{table}[h!]
\caption{Reference literature that considered GAN-based approaches to handle class imbalanced problems.}
\label{tab:compare}
\resizebox{\textwidth}{!}{%
\begin{tabular}{@{}lllllll@{}}
\toprule
\textbf{Reference} & \textbf{Contributions} & \textbf{Algorithms} & \textbf{Dataset} & \textbf{Performance} & \textbf{\begin{tabular}[c]{@{}l@{}}Misclassification \\ Evaluation\end{tabular}} & \textbf{\begin{tabular}[c]{@{}l@{}}Algorithm \\ Complexity\end{tabular}} \\ \midrule
\begin{tabular}[c]{@{}l@{}}Li et al. \\ (2022)~\cite{li2022hybrid}\end{tabular} & \begin{tabular}[c]{@{}l@{}}Hybrid \\ method\end{tabular} & \begin{tabular}[c]{@{}l@{}}Cluster-\\ Borderline \\ SMOTE\end{tabular} & \begin{tabular}[c]{@{}l@{}}Rock \\ Groutability\end{tabular} & \begin{tabular}[c]{@{}l@{}}Improved \\ AUC and \\ F1-Score\end{tabular} & \begin{tabular}[c]{@{}l@{}}Confusion Matrix, \\ ROC Curve, \\ AUC, \\ F1-Score\end{tabular} & - \\
\begin{tabular}[c]{@{}l@{}}Ning et \\ al. (2021)~\cite{ning2021novel}\end{tabular} & \begin{tabular}[c]{@{}l@{}}Hybrid \\ method\end{tabular} & \begin{tabular}[c]{@{}l@{}}SMOTE with \\ Tomek Links\end{tabular} & \begin{tabular}[c]{@{}l@{}}Glutarylation \\ Sites\end{tabular} & \begin{tabular}[c]{@{}l@{}}Enhanced \\ performance \\ of the classifier\end{tabular} & \begin{tabular}[c]{@{}l@{}}Confusion Matrix, \\ ROC Curve, \\ Precision, Recall, \\ F1-Score\end{tabular} & - \\
\begin{tabular}[c]{@{}l@{}}Zhang et \\ al. (2020)~\cite{zhang2020network}\end{tabular} & \begin{tabular}[c]{@{}l@{}}Hybrid \\ method\end{tabular} & \begin{tabular}[c]{@{}l@{}}ReliefF with \\ Borderline-\\ SMOTE\end{tabular} & \begin{tabular}[c]{@{}l@{}}Intrusion \\ Detection\end{tabular} & \begin{tabular}[c]{@{}l@{}}Improved \\ performance \\ of the classifier\end{tabular} & \begin{tabular}[c]{@{}l@{}}Confusion Matrix, \\ ROC Curve, \\ Precision, Recall, \\ F1-Score\end{tabular} & - \\
\begin{tabular}[c]{@{}l@{}}Sun et \\ al. (2020)~\cite{sun2020class}\end{tabular} & \begin{tabular}[c]{@{}l@{}}Ensemble \\ method\end{tabular} & \begin{tabular}[c]{@{}l@{}}Adaboost-SVM \\ with \\ SMOTE\end{tabular} & \begin{tabular}[c]{@{}l@{}}Chinese \\ Listed Cos.\end{tabular} & \begin{tabular}[c]{@{}l@{}}Improved \\ performance \\ of the Classifier\end{tabular} & \begin{tabular}[c]{@{}l@{}}Confusion Matrix, \\ ROC Curve, \\ Precision, Recall, \\ F1-Score\end{tabular} & - \\
\begin{tabular}[c]{@{}l@{}}Liang et \\ al. (2020)~\cite{liang2020lr}\end{tabular} & \begin{tabular}[c]{@{}l@{}}Hybrid \\ method\end{tabular} & \begin{tabular}[c]{@{}l@{}}K-means with \\ SVM\end{tabular} & - & \begin{tabular}[c]{@{}l@{}}Proposed models \\ can \\ generate \\ samples without \\ considering outliers\end{tabular} & - & - \\
\begin{tabular}[c]{@{}l@{}}Ali-Gombe et \\ al. (2019)~\cite{ali2019mfc}\end{tabular} & \begin{tabular}[c]{@{}l@{}}GAN-based \\ method\end{tabular} & MFC-GAN & \begin{tabular}[c]{@{}l@{}}Synthetic \\ Data\end{tabular} & \begin{tabular}[c]{@{}l@{}}Improved \\ classification \\ performance\end{tabular} & \begin{tabular}[c]{@{}l@{}}Confusion Matrix, \\ Precision, Recall, \\ F1-Score\end{tabular} & High \\
\begin{tabular}[c]{@{}l@{}}Kim et \\ al. (2020)~\cite{kim2020gan}\end{tabular} & \begin{tabular}[c]{@{}l@{}}GAN-based \\ method\end{tabular} & \begin{tabular}[c]{@{}l@{}}GAN-based \\ approach\end{tabular} & \begin{tabular}[c]{@{}l@{}}Anomaly \\ Detection\end{tabular} & \begin{tabular}[c]{@{}l@{}}Improved \\ detection \\ accuracy\end{tabular} & \begin{tabular}[c]{@{}l@{}}Confusion Matrix, \\ ROC Curve, Precision, \\ Recall, F1-Score\end{tabular} & High \\
\begin{tabular}[c]{@{}l@{}}Rajabi et \\ al. (2022)~\cite{rajabi2022tabfairgan}\end{tabular} & \begin{tabular}[c]{@{}l@{}}GAN-based \\ method\end{tabular} & TabFairGAN & \begin{tabular}[c]{@{}l@{}}Tabular \\ Data\end{tabular} & \begin{tabular}[c]{@{}l@{}}Promising \\ performance on \\ multiple \\ benchmark \\ datasets\end{tabular} & \begin{tabular}[c]{@{}l@{}}Confusion Matrix, \\ ROC Curve, Accuracy, \\ F1-Score\end{tabular} & High \\
\begin{tabular}[c]{@{}l@{}}Engelmann and \\ Lessmann (2021)~\cite{engelmann2021conditional}\end{tabular} & \begin{tabular}[c]{@{}l@{}}GAN-based \\ method\end{tabular} & cWGAN & \begin{tabular}[c]{@{}l@{}}Tabular \\ Data\end{tabular} & \begin{tabular}[c]{@{}l@{}}Improved Logistic \\ Regression \\ ranking\end{tabular} & \begin{tabular}[c]{@{}l@{}}Confusion Matrix, \\ ROC Curve, Precision, \\ Recall, F1-Score\end{tabular} & High \\
\begin{tabular}[c]{@{}l@{}}Jo et \\ al. (2022)~\cite{jo2022obgan}\end{tabular} & \begin{tabular}[c]{@{}l@{}}GAN-based \\ method\end{tabular} & OBGAN & \begin{tabular}[c]{@{}l@{}}Imbalanced \\ Datasets\end{tabular} & \begin{tabular}[c]{@{}l@{}}Highest \\ Recall and F1-Score \\ among the \\ tested \\ techniques\end{tabular} & \begin{tabular}[c]{@{}l@{}}Confusion Matrix, \\ ROC Curve,  Precision, \\ Recall, F1-Score\end{tabular} & High \\ \bottomrule
\end{tabular}%
}
\end{table}

\section{Methodology}\label{met}
In this section, we discuss in detail the algorithms such as SMOTE, Borderline-SMOTE, GAN, and our proposed approach, BSGAN.
\subsection{SMOTE}
SMOTE is one of the most widely used oversampling techniques in ML domains, proposed by Chawla~\citep{chawla2002smote}. The SMOTE algorithm has the following input parameters that can be controlled and changed: K as the number of nearest neighbors (default value, k = 5), and oversampling percentage parameters (default value 100\%).

In SMOTE, a random sample is initially drawn from the minor class. Then k-nearest neighbors are identified to observe the random samples. After that, one of the neighbors is taken to identify the vector between the instant data point and the selected neighbors. The newly found vector is multiplied by the random number between 0 to 1 to generate new instances from the initial minor instance on the line. Then SMOTE continues the same process with other minor samples until it reaches the percentage value assigned by the user. Algorithm~\ref{SMOTE} displays the pseudocode of SMOTE, where the appropriate function is introduced for each step of SMOTE process. From the algorithm, it can be observed that it takes as input the number of instances in the minority class (P), the percentage of synthetic samples to be generated (S), and the number of nearest neighbors to consider (K). Using a randomly generated gap value, the algorithm generates synthetic samples by interpolating between a selected instance and one of its nearest neighbors. The number of synthetic samples to be generated equals P times S/100. To achieve this, SMOTE first finds the K nearest neighbors for each instance in the minority class and saves their indices in an array. The algorithm then repeats this process until the desired number of synthetic samples has been generated. By creating synthetic samples, SMOTE can improve the accuracy of machine learning models in predicting the minority class, thereby making them more effective in real-world applications.
\begin{algorithm}
\caption{SMOTE}\label{SMOTE}
\begin{algorithmic}
\STATE Input: P number of minor class sample; $S\%$ of synthetic to be generated; K Number of nearest neighbors
\STATE Output: $N_s=(S/100)*P$ synthetic samples
\STATE 1. \textbf{Create function ComputKNN ($i \gets 1 to P, P_i, P_j$)}
\FOR{$i \gets 1 to P$}
\STATE Compute K nearest neighbors of each minor instance $P_i$ and other minor instance $P_j$.
\STATE Save the indices in the nnaray.
\STATE Populate ($N_s, i, nnarray$) to generate new instance.
\ENDFOR
\STATE $N_S=(S/100)*P$
\WHILE{$N_s \neq 0$}
\STATE \textbf{Create function GenerateS ($P_i$, $P_j$)}
\STATE {Choose a random number between 1 and K, call it nn.
\FOR{$att \gets 1$ to numattrs}
\STATE dif= $P_i [nnarray[nn]][attr]-P_j[i][attr]$
\STATE $gap = random number between 0 and 1$
\STATE $Synthetic[newindex[attr] = P_i[i][attr]+gap*dif$

\ENDFOR
\STATE newindex = newindex + 1
 \STATE $N_s = N_s-1$}
\ENDWHILE
\STATE 4. Return ($*End of Populate.*$)
\STATE End of Pseudo-Code.
\end{algorithmic}
\end{algorithm}

    

As mentioned earlier, SMOTE generates randomly new samples on the datasets, which increases the noise in the major class area, or within the safe minor region far from the borderline area and overfitting it, therefore not efficiently increasing the classification accuracy in order to classify the minor samples. As an effect, SMOTE has several derivatives, such as Borderline-SMOTE, SMOTEBOOST, Safe-level-SMOTE, and others, which were introduced to limit or reduce these problems. This research primarily focuses on utilizing and modifying the Borderline-SMOTE to overcome the existing limitations mentioned in section~\ref{rw}.
\subsection{Borderline-SMOTE}
Borderline-SMOTE is a popular extension of the SMOTE that is designed to handle imbalanced datasets in ML domains. Borderline-SMOTE was proposed to address some of the limitations of SMOTE for imbalanced dataset classification. Unlike SMOTE, which randomly interpolates between minority samples, Borderline-SMOTE specifically focuses on synthesizing new samples along the borderline between the minority and majority classes. This approach helps to improve the class balance in the dataset and prevent the model from overfitting to the majority class~\citep{han2005borderline}. The Borderline-SMOTE algorithm extends the traditional SMOTE by differentiating between minority samples by utilizing the M' number of majority instances within the M-Nearest Neighbors (MNN) of a given minority instance $P_i$. The default value of M is set to 5. The minority instance is considered safe if the number of majority instances within its MNN is within the range of 0 to M/2. On the other hand, if all of the MNN of a minority instance consist of majority instances, with $M'=M$, the instance is considered to be noise and is eliminated from the computation function to reduce oversampling near the border. Finally, a minority instance is considered a danger instance P' if the number of majority instances within its MNN falls within the range of M/2 to M. After that, Borderline-SMOTE measures KNN between borderline instance and minor instances and generates a new instance using the following equations~\citep{fernandez2018smote, han2005borderline}:
\begin{equation}
    \mbox{New instance} = P'_i + \mbox{gap}* (\mbox{distance} (P'_i,P_j))
\end{equation}

Where $P'_i$ is the borderline minor instance, $P_j$ is the randomly chosen KNN minor instance, and a gap is a random number between 0 and 1. Algorithm~\ref{bsmote} displays the pseudocode of B-SMOTE. One of the potential drawbacks of B-SMOTE is that it focuses on the borderline region; therefore, widening the region might confuse the classifier.
\begin{algorithm}
\caption{Pseudocode for Borderline-SMOTE}\label{bsmote}
\begin{algorithmic}[]
\STATE Input: P number of minor sample; s\% of synthetic to generate; M number of nearest neighbors to create the borderline subset; k Number of nearest neighbors
\STATE \textbf{Output:} $(s/100)^*P'$ synthetic samples
\STATE 1. \textit{\textbf{Creating function MinDanger ()}}
\FOR{$i \gets 1 to P$}
\STATE Compute M nearest neighbors of each minor instance and other instances from the dataset,
\STATE Check the number of Major instance M' within the Mnn
\IF{M/2<M'<M}
\STATE Add instance P to borderlines subset P'
\ENDIF
\ENDFOR
\STATE 2. \textit{\textbf{ComputeKNN ($i \gets 1 to P', P_i, P_j$)}}
\STATE 3. $N_s=(S/100)*P'$
\WHILE{$N_s \neq 0$}
\STATE 4. $GenerateS (P_i', P_j)$
\STATE $N_s = N_s -1 $
\ENDWHILE
\STATE 5. Return ($*$End of Populate.$*$)
\STATE End of Pseudo-Code.
\end{algorithmic}
\end{algorithm}
\subsection{GAN}
GAN is a class of ML frameworks that contains two Neural Networks (NN). The goal of this framework is to train both networks simultaneously and improve their performance while reducing their loss function as well. Following true data distribution, a new sample is generated with the same statistics as the training set~\citep{sharma2022smotified}. The pseudocode for the GAN algorithm is presented in Algorithm~\ref{gan}, where Stochastic Gradient Descent (SGD) and weights are defined functions that determine mini-batch gradient or any other variant such as Adaptive Momentum (ADAM) or Root Mean Square Propagation (RMSprop) and update the weights respectively~\citep{nugroho2022performance, hameed2016back, ketkar2017stochastic}. Once the algorithm terminates, ‘good’ fake samples are collected with accumulateFakeEx based on classification accuracy.

GAN typically contains two NN: generator ($G$) and discriminator ($D$). The goal of the G is to create fake samples that look almost real. A random noise between $0$ and $1$ is used initially to create fake samples. On the other hand, $D$ is trained with the real sample from the dataset. A random sample created by G is then passed to $D$ so that $D$ can distinguish between the real and the fake samples. The goal of the $G$ is to fool the $D$ by creating fake samples which look like reals. Conversely,  the goal of the $D$ is not to get fooled by $G$. During this process, both $D$ and $G$ optimize their learning process. The loss function for $D$ can be calculated as follows~\cite{goodfellow2016deep}:
\begin{equation}
     \max_D\mathds{E}_x[logD(x)]+\mathds{E}_z[log(1-D(G(z)))] 
\end{equation}
Where the notation $D(x)$ represents the probability distribution obtained from a real data sample $x$, while $D(G(z))$ refers to the probability distribution produced by a generated sample $z$.

The loss function of $G$ can be calculated as follows:
\begin{equation}
     \min_G-\mathds{E}_z[logD(G(z))]
 \end{equation}

\begin{algorithm}
\caption{Pseudocode for GAN}\label{gan}
\begin{algorithmic}[5]
\STATE // Input: training data set examples x
and noise samples z from
appropriate random number generator. An optional parameter
can be the size nfake of fake sample
needed.

// initialize parameters

// mi is the minibatch indices for i
th
index and T is the total iterations.
\STATE GAN ($x,z, n_{fake}$)

\FOR{t=1:T} 

\STATE //Generally, step size S is 1

\STATE // subscript d and g refer to
discriminator and generator
entity respectively
\FOR{$s=1:S$}
\STATE $g_d \gets$
\STATE $SGD(- \log D(x) - \log(1 - D(G(z)), W_d , m_i)$
\STATE $W_d \gets weights(g_d , W_d )$
\STATE $W_g \gets weights(g_g, W_g)$

\ENDFOR
\ENDFOR
\STATE $x' \gets$ accumulateFakeEx ($Model_d(W_d,x,z), 
Model_g (W_g, x,z), n_{fake})$
\RETURN $x'$
\end{algorithmic}
\end{algorithm}
\subsection{Proposed BSGAN}
\begin{figure*}[h]
    \centering
    \includegraphics[width=\textwidth]{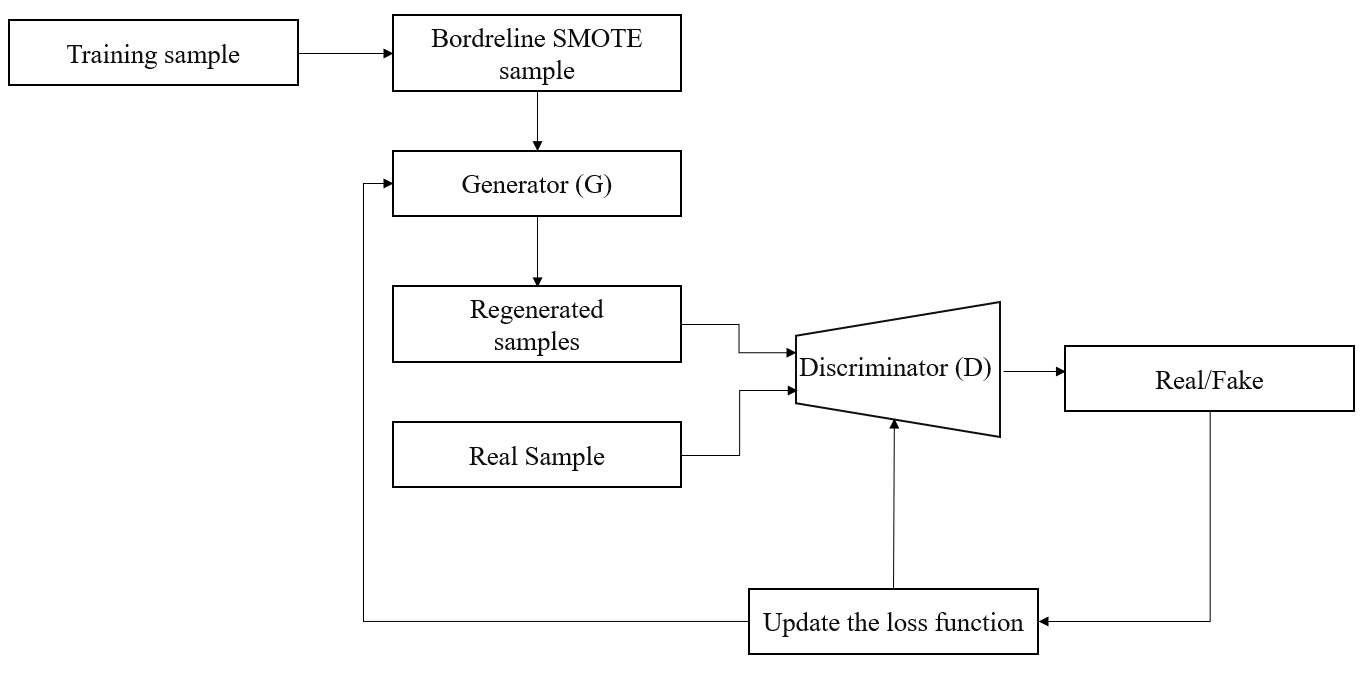}
    \caption{Flow diagram of Proposed Borderline-SMOTE and Generative Adversarial Networks (BSGAN) models. }
    \label{fig:fig2}
\end{figure*}

Our proposed approach combined borderline SMOTE and naïve GAN to handle class imbalance problems. The borderline SMOTE starts by classifying the minor class observations. If all the neighbors are close to the major class, it classifies any minor samples as a noise point. Further, it classifies a few points as border points with major and minor classes close to the neighborhood and resamples from them.
 In our proposed BSGAN, we modified the loss function of GAN and combined them with the borderline SMOTE algorithms. Here, instead of random noise for the $G$, we are passing a sample created by borderline SMOTE. The updated loss for the $D$ can be expressed as follows:
 
 \begin{equation}
   \max_D\mathds{E}_{x^*}[logD(x^*|x)]+\mathds{E}_u[log(1-D(G(u)))] 
\end{equation}
The updated loss for the $G$ can be expressed as follows:
\begin{equation}
     \min_G-\mathds{E}_z[logD(G(u))]
\end{equation}
Where, $x^*$ = training sample of minor class\\
$U$ = oversampled data generated by borderline SMOTE.\\
Figure~\ref{fig:fig2} demonstrates the overall flow diagram of the proposed BSGAN algorithms.

The pseudocode of the proposed BSGAN is described in Algorithm~\ref{bsgan}. As illustrated in Algorithm~\ref{bsgan}, there are two sections of BS-GAN. The first one replaces the random number sample from the sample generated by borderline-SMOTE. The second section continues with the process of GAN using the new samples from the B-SMOTE. Algorithm 4 also shows this whole procedure in two steps. In-Line (1) calls the BS-SMOTE function in Algorithm 2, and then Line (2) calls the modified GAN function given in Algorithm 3. However, this time the generated sample u is used instead of random noise z.
\begin{algorithm}
\caption{: Pseudocode for BSGAN}\label{bsgan}
Step 1 $\rightarrow$ Input: minor samples $X^*$ from the training data $x$ of size $N$ that requires $N$ -- $n$ over-samples;\\
Step 2 $\rightarrow$ User-defined parameter $k$ for K-nearest neighbors\\
Step 3 $\rightarrow$ Execute Borderline-SMOTE given in Algorithm 1 then GAN given in Algorithm 2\\
1 $u \leftarrow$ call Algorithm 1 ($x^*,k$)// generate over-sampled minor examples $u$.\\
2 $u \leftarrow$ call Algorithm 2 ($x^*$,$u$,$N$ - $n$).\\
\end{algorithm}

\subsection{Proposed Neural Network}
A neural network model is used to train and test the model on a different dataset. Parameters such as batch size, number of epochs, learning rate, and the hidden layer are tuned manually by trial and error process. Table~\ref{tab:tab1} presents the details of the optimized parameters obtained throughout the experiment to achieve the best experimental outcomes for the discriminator, generator, and neural network. The number of epochs varies for each dataset as each dataset differs due to different features and sample sizes.

\begin{table*}[h]
\caption{Parameter settings used to develop discriminator, generator, and neural network.}
\label{tab:tab1}
\resizebox{0.8\textwidth}{!}{%
\begin{tabular}{llll}
\hline
\textbf{Parameters} & \textbf{Discriminator} & \textbf{Generator} & \textbf{Neural Network} \\ \hline
Number of hidden layer & 4 & 3 & 3 \\
Number of neurons & 64,128,256,512 & 512, 256,128 & 256, 128,1 \\
Batch size & 32 & 32 & 32 \\
Learning rate & 0.00001 & 0.00001 & 0.00001 \\
Optimizer & Adam & Adam & Adam \\
Loss function & Binary cross entropy & Binary cross entropy & Binary cross entropy \\
Activation function & ReLU & ReLU & ReLU \& Sigmoid \\ \hline
\end{tabular}%
}
\end{table*}

 
   
\section{Performance Evaluation}\label{res}
\subsection{Datasets}
We evaluate and compare our model on four distinct highly imbalanced datasets--Ecoli, Yeast, Wine quality, and Abalone--that feature class imbalance, as shown in Table~\ref{tab:tab2}. The datasets were primarily adopted from the UCI machine learning repository, which has been used by researchers and practitioners to evaluate the model performance for CIPs. Some datasets, such as Wine quality and Ecoli, are highly imbalanced and contain only 2.74\% and 5.97\% minority classes.


\begin{table*}[h]
\centering
\caption{Characteristics of imbalanced dataset utilized for the experiment.}
\label{tab:tab2}
\resizebox{.8\textwidth}{!}{%
\begin{tabular}{lllllll}
\toprule
\textbf{Dataset} & \textbf{\# of sample} & \textbf{\begin{tabular}[c]{@{}l@{}}Minor \\ sample\end{tabular}} & \textbf{\begin{tabular}[c]{@{}l@{}}Major \\ sample\end{tabular}} & \textbf{\begin{tabular}[c]{@{}l@{}}Total \\ features\end{tabular}} & \textbf{\begin{tabular}[c]{@{}l@{}}Minority \\ class(\%)\end{tabular}} & \textbf{Description} \\ \midrule
\textbf{\textbf{Ecoli}} & 335 & 20 & 315 & 7 & 5.97 & Protein localization \\
\textbf{\textbf{Yeast}} & 513 & 51 & 462 & 8 & 9.94 & \begin{tabular}[c]{@{}l@{}}Predicting protein \\ localization cite.\end{tabular} \\
\textbf{\textbf{Winequality}} & 655 & 18 & 637 & 10 & 2.74 & \begin{tabular}[c]{@{}l@{}}Classify the \\ wine quality\end{tabular} \\
\textbf{\textbf{Abalone}} & 4177 & 840 & 3337 & 8 & 20.1 & \begin{tabular}[c]{@{}l@{}}Predict the age \\ of abalone\end{tabular} \\ \bottomrule
\end{tabular}%
}
\end{table*}

   
\subsection{Experimental Setup}
 An office-grade laptop with standard specifications (Windows 10, Intel Core I7-7500U, and 16 GB of RAM) is used to conduct the whole experiment. The empirical experiment was carried out five times, and the final results are presented by averaging all five outcomes. Initially, the dataset is split into the following ratios— trainset/test set: 80/20. The experimental evaluation results are presented in terms of accuracy, precision, recall, F1-score, and AUC-ROC score.
 
\textbf{\textit{Accuracy}:} The accuracy reflects the total number of instances successfully identified among all instances. The following formula can be used to calculate accuracy.
\begin{equation}\label{acc}
    Accuracy = \frac{T_{p}+ T_{N}}{T_{p}+T_{N}+F_{p}+F_{N}}
\end{equation}
\textbf{\textit{Precision}}
Precision is defined as the percentage of accurately anticipated positive observations to all expected positive observations.
\begin{equation}
    Precision= \frac{T_p}{T_p+F_p}
\end{equation}
\textbf{\textit{Recall:}}
The recall is the percentage of total relevant results that the algorithm correctly detects.
\begin{equation}
   Recall= \frac{T_p}{T_n+F_p}
\end{equation}

\textbf{\textit{F1-score:}} 
The F1-score is the mean of accuracy and recall in a harmonic manner. The highest f score is 1, indicating perfect precision and recall score.
\begin{equation}\label{f1}
 F1- score=2\times\frac{\textrm{Precision}\times\textrm{ Recall}}{\textrm{Precision+Recall}}
\end{equation}
\textit{\textbf{Area under curve (AUC):}}
The area under the curve illustrates how the models behave in various conditions. The AUC can be measured using the following formula:
\begin{equation}
    AUC= \frac{\sum R_i(I_p)- I_p((I_p +1)/2}{I_p+I_n}
\end{equation}

Where, $l_p$  and $l_n$ denotes positive and negative data samples and $R_i$ is the rating of the $i^{th}$ positive samples.\\
True Positive ($T_p$)= Positive samples classified as \mbox{Positive}\\
False Positive ($F_p$)= Negative samples classified as \mbox{Positive}\\ 
True Negative ($T_n$)= Negative samples classified as \mbox{Negative}\\
False Negative ($F_n$)= Positive samples classified as \mbox{Negative}\\

\begin{equation}\label{inta}
\text{Interclass distance} = \frac{\mu_1 - \mu_2}{\sqrt{\frac{1}{n_1} + \frac{1}{n_2}}}
\end{equation}
This assumes that there are two classes with means $\mu_1$ and $\mu_2$, and sample sizes of $n_1$ and $n_2$, respectively.

\subsection{Results}
The overall performance for data with and without oversampling was measured using equations~\ref{acc}--\ref{f1} and presented in Table~\ref{tab:result}. The best results are highlighted with bold fonts. From the table, it can be seen that the Proposed BSGAN outperformed all of the techniques across all measures in all datasets. However, on the Wine quality dataset, GAN and BSGAN both demonstrated similar performance on the train set by achieving an accuracy of 99.17\%. The highest F1-score was achieved using BSGAN (0.9783) on the Yeast dataset. The lowest F1-score was achieved on the Abalone dataset when tested without oversampling techniques (0.9041). The highest recall score of 1.0 was achieved on the Winequality dataset using BSGAN. On the other hand, the lowest recall score of 0.9055 was achieved on the Abalone dataset when the dataset was tested without oversampling techniques. A maximum precision score of 0.9768 was achieved on the Ecoli dataset using BSGAN, while the lowest precision score of 0.9036 was observed on the Abalone dataset.

\begin{table}[h]
\centering
\caption{Performance evaluation of different Oversampling techniques used in this study on highly imbalanced benchmark datasets.}
\label{tab:result}
\resizebox{0.8\textwidth}{!}{%
\begin{tabular}{lllllll}
\hline
\multirow{2}{*}{\textbf{Dataset}} & \textbf{Oversampling} & \multicolumn{1}{c}{\textbf{Train}} & \multicolumn{4}{c}{\textbf{Test}} \\ \cline{2-7} 
 & \textbf{Strategy} & \textbf{accuracy} & \textbf{accuracy} & \textbf{Precision} & \textbf{Recall} & \textbf{F1-score} \\ \hline
\multirow{4}{*}{\textbf{Ecoli}} & \begin{tabular}[c]{@{}l@{}}Without-\\ oversampling\end{tabular} & 93.22\% & 91.67\% & 0.9167 & 0.9167 & 0.9095 \\
 & \begin{tabular}[c]{@{}l@{}}Borderline-\\ SMOTE\end{tabular} & 98.84\% & 95.11\% & 0.9661 & 0.9523 & 0.9572 \\
 & GAN & 98.33\% & 97.61\% & 0.9767 & 0.9761 & 0.9703 \\
 & BSGAN & 99.29\% & \textbf{97.85\%} & \textbf{0.9786} & \textbf{0.9785} & \textbf{0.9783} \\
\multirow{4}{*}{\textbf{Yeast}} & \begin{tabular}[c]{@{}l@{}}Without-\\ oversampling\end{tabular} & 92.61\% & 90.72\% & 0.9043 & 0.9072 & 0.9042 \\
 & \begin{tabular}[c]{@{}l@{}}Borderline-\\ SMOTE\end{tabular} & 87.89\% & 92.32\% & 0.9347 & 0.9232 & 0.9274 \\
 & GAN & 97.11\% & 94.18\% & 0.9396 & 0.9418 & 0.9351 \\
 & BSGAN & 97.17\% & \textbf{94.65\%} & \textbf{0.9441} & 0.9465 & 0.9412 \\
\multirow{4}{*}{\textbf{Wine quality}} & \begin{tabular}[c]{@{}l@{}}Without-\\ oversampling\end{tabular} & 98.37\% & 93.90\% & 0.9390 & \textbf{1.0} & \textbf{0.9685} \\
 & \begin{tabular}[c]{@{}l@{}}Borderline-\\ SMOTE\end{tabular} & 99.03\% & 92.68\% & 0.9068 & 0.9268 & 0.9150 \\
 & GAN & 99.17\% & 93.84\% & 0.9332 & 0.9932 & 0.9623 \\
 & BSGAN & 99.17\% & \textbf{93.90\%} & \textbf{0.9390} & \textbf{1.0} & \textbf{0.9685} \\
\multirow{4}{*}{\textbf{Abalone}} & \begin{tabular}[c]{@{}l@{}}Without-\\ oversampling\end{tabular} & 90.37\% & 90.55\% & 0.9036 & 0.9055 & 0.9041 \\
 & \begin{tabular}[c]{@{}l@{}}Borderline-\\ SMOTE\end{tabular} & 87.17\% & 84.21\% & 0.8945 & 0.8421 & 0.8539 \\
 & GAN & 94.09\% & 90.54\% & 0.9032 & 0.9054 & 0.9037 \\
 & BSGAN & 94.18\% & \textbf{90.64\%} & \textbf{0.9049} & \textbf{0.9064} & \textbf{0.9052} \\ \bottomrule
\end{tabular}%
}
\end{table}

\begin{figure}[h]
    \centering
    \includegraphics[width=.5\textwidth]{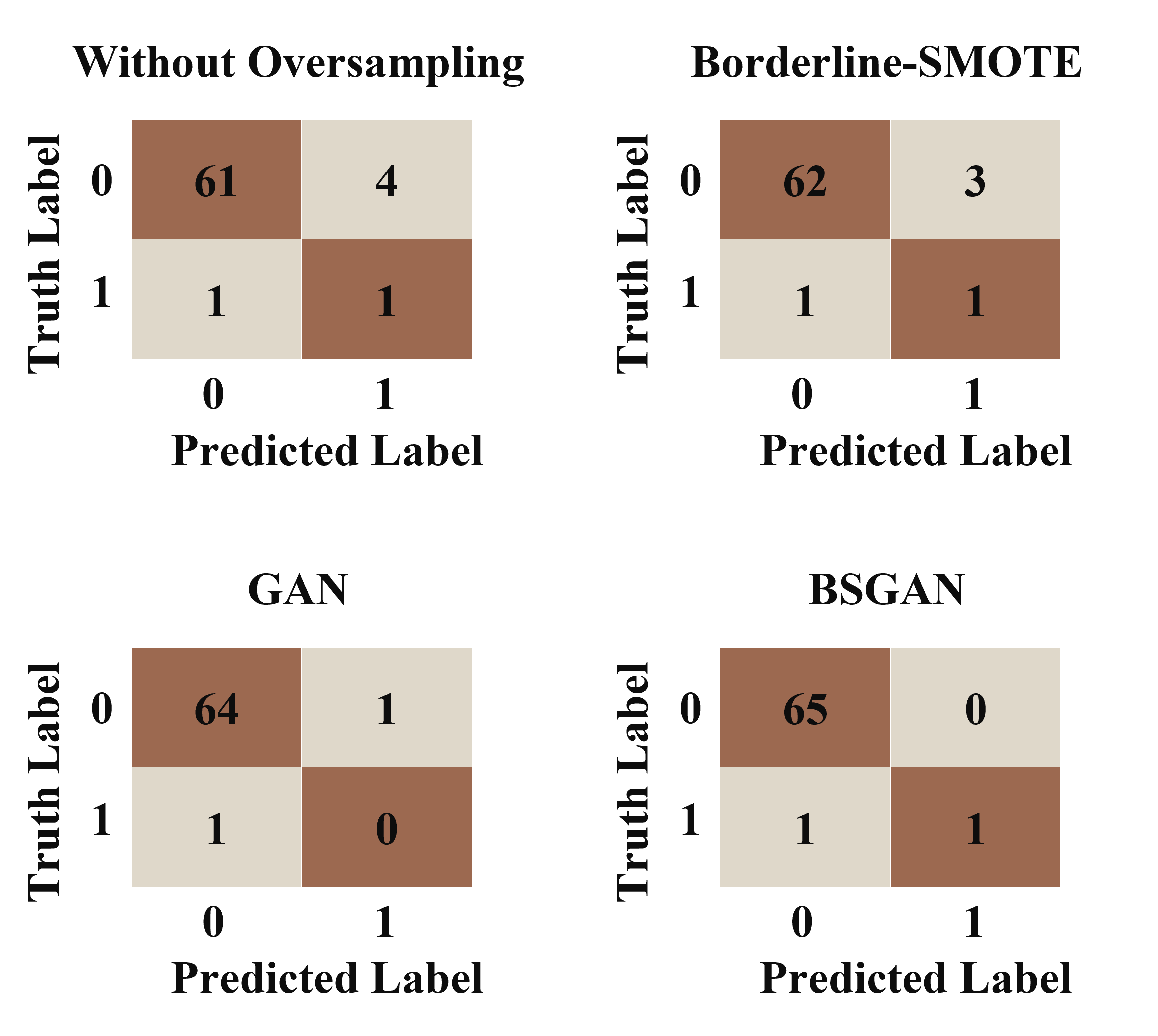}
    \caption{Performance measurement of without and with oversampling techniques on Ecoli test dataset using confusion matrices.}
    \label{fig:cf1}
\end{figure}

The confusion matrix was calculated on the test set to simplify the understanding of the performance of different oversampling techniques on different imbalanced datasets. Figure~\ref{fig:cf1} displays the confusion matrix for different sampling techniques on a given Ecoli test dataset. On the Ecoli dataset, maximum misclassification occurred for the dataset without oversampling techniques, up to 7.46\% (5 samples). On the other hand, minimum misclassification occurred for BSGAN, up to 1.49\% (only one sample).

\begin{figure}[h!]
    \centering
    \includegraphics[width=.5\textwidth]{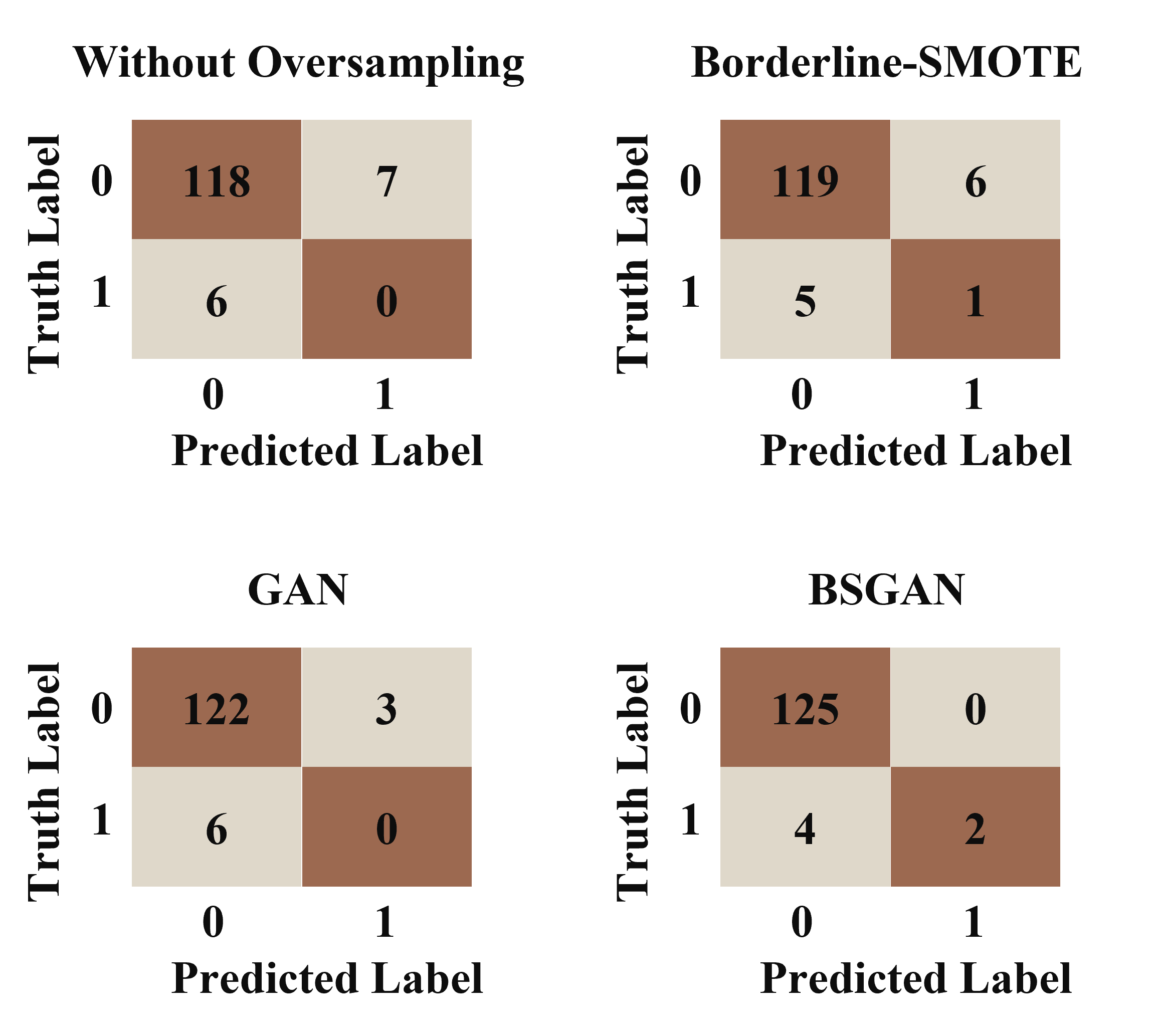}
    \caption{Performance measurement of without and with oversampling techniques on Winequality test dataset using confusion matrices.}
    \label{fig:fig4}
\end{figure}

Figure~\ref{fig:fig4} displays the confusion matrix for different sampling techniques on a Wine quality test dataset. The figure shows that the NN model performance on the Wine quality dataset without oversampling demonstrated the worst classification by misclassifying 13 out of 131 samples (9.9\%). In comparison, BSGAN showed the best performance by misclassifying only 4 out of 131 samples (3.05\%).

\begin{figure}[h!]
    \centering
    \includegraphics[width=.5\textwidth]{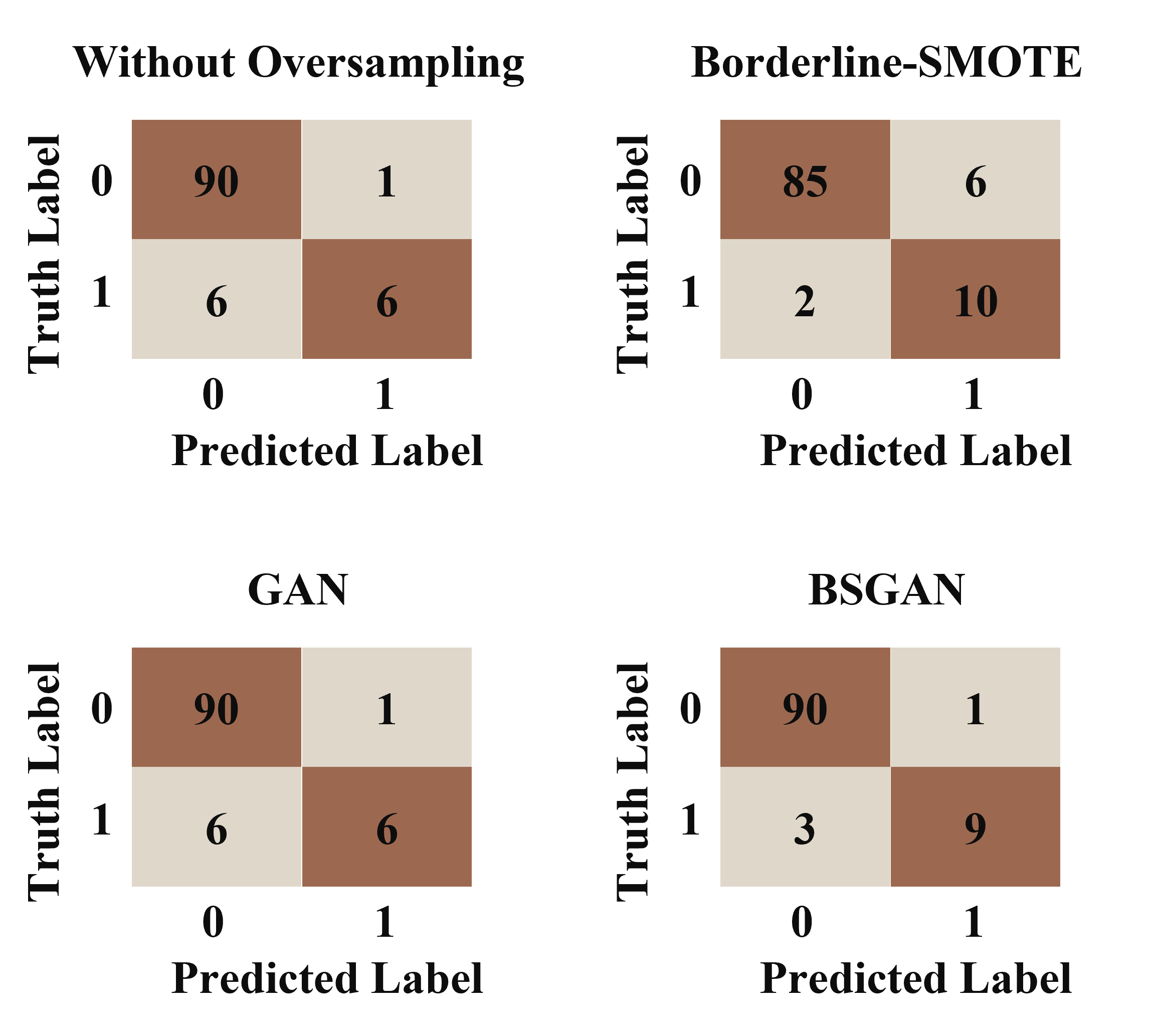}
    \caption{Performance measurement of without and with oversampling techniques on the Yeast test dataset using confusion matrices.}
    \label{fig:fig5}
\end{figure}

Figure~\ref{fig:fig5} displays the confusion matrix for different sampling techniques on a given Yeast test dataset. From the figure, it can be observed that NN model performance on the yeast dataset Borderline-SMOTE demonstrated the worst performance by misclassifying 8 out of 131 samples (7.77\%), while BSGAN showed the best performance by misclassifying only four samples (3.88\%).

\begin{figure}[h!]
    \centering
    \includegraphics[width=.5\textwidth]{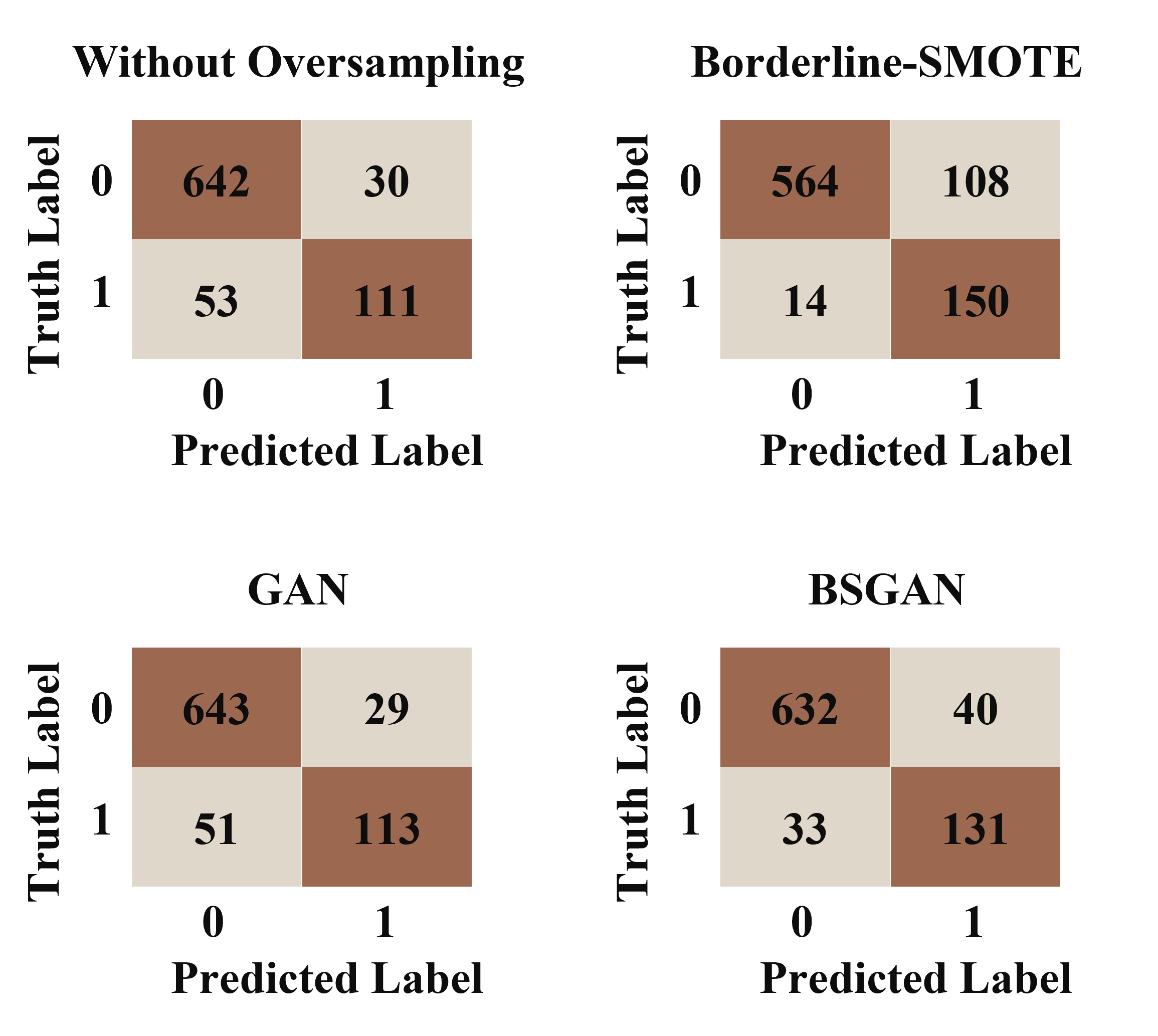}
    \caption{Performance measurement of without and with oversampling techniques on Abalone test dataset.}
    \label{fig:fig6}
\end{figure}

Figure~\ref{fig:fig6} illustrates the confusion matrix for different sampling techniques on a given Abalone test dataset. From the figure, it can be observed that NN model performance on the Abalone dataset Borderline-SMOTE demonstrated the worst performance by misclassifying 122 out of 836 samples (14.59\%), while BSGAN showed the best performance by misclassifying 73 samples (8.73\%).

To understand the data distribution after expanding the dataset using, different oversampling techniques have been measured using equation~\ref{inta}. The closer the inter-class distance between the dataset and the expanded data, the better the classification effect, ultimately demonstrating better Gaussian distributions. From Table~\ref{tab:tab4}, it can be observed that the interclass distance between the BSGAN and the dataset without oversampling is the closest compared to any other oversampling techniques used in this study. On the Abalone dataset, Borderline-SMOTE also demonstrates the closest inter-class distance with original datasets. Unfortunately, data expansion after applying GAN shows the worst performance on three out of four imbalanced datasets— Ecoli, Wine quality, and Abalone.

\begin{table}[h!]

\caption{The inter-class distance between the original datasets and the datasets after the expansion using different oversampling techniques.}
\centering
\label{tab:tab4}
\resizebox{.55\textwidth}{!}{%
\begin{tabular}{lllll}
\hline
\textbf{Dataset} & \textbf{WS} & \textbf{S} & \textbf{GBO} & \textbf{SSG} \\ \hline
\textbf{Ecoli} & 0.1650 & 0.1352 & 0.0893 & 0.150 \\
\textbf{Yeast} & 0.093 & 0.079 & 0.083 & 0.10 \\
\textbf{Wine quality} & 0.1541 & 0.1531 & .0871 & 0.158 \\
\textbf{Abalone} & 0.2633 & 0.25 & 0.1856 & 0.25 \\ \hline
\end{tabular}%
}
\end{table}

\section{Discussion}\label{dis}
As a means of comparing our results with those available in the literature, Table~\ref{tab:tab5} contrasts the performance of our proposed methods on Yeast datasets in terms of accuracy, precision, recall, and F1-score. The table shows that BSGAN outperformed all of the referenced literature across all measures except the performance of accuracy. While Jadhav et al. (2020) achieved the highest accuracy (98.42\%), their precision score is relatively deficient, and their F1-score is 0, which hinders a direct comparison of all reported performance measures.

\begin{table}[h!]
\centering
\caption{Comparison with previous studies on Yeast datasets.}
\label{tab:tab5}
\resizebox{.8\textwidth}{!}{%
\begin{tabular}{llllll}
\toprule
\textbf{Author} & \textbf{Techniques} & \textbf{Accuracy} & \textbf{Precision} & \textbf{Recall} & \textbf{F1-score} \\ \midrule
\textbf{~\cite{sharma2022smotified}} & SMOTified-GAN & 96.11\% & 0.91 & 0.83 & 0.873 \\
\textbf{~\cite{siddappa2019adaptive}} & LMDL & 56.87\% & .57 & .57 & .55 \\
\textbf{~\cite{karia2019gensample}} & GenSample & 70\% & 0.47 & 0.50 & 0.48 \\
\textbf{~\cite{jo2022obgan}} & OBGAN & - & - & 0.6135 & 0.5556 \\
\textbf{~\cite{jadhav2020novel}} & svmradial & 98.42\% & 0.8 & - & 0 \\
\textbf{Our study} & BSGAN & 97.17\% & 0.9441 & 0.9465 & 0.9412 \\ \bottomrule
\end{tabular}%
}
\end{table}

On Ecloi datasets, our proposed BSGAN demonstrates consistent performance and outperformed all of the referenced literature in terms of accuracy by achieving an accuracy of 99.29\%. Sharma et al. (2022) claimed 100\% precision, recall, and F1-score while the accuracy is only 90.75\%. Therefore, there is some discrepancy in the results reported by the authors. 

\begin{table}[h!]
\centering
\caption{Comparison with the previous study on Ecoli datasets.}
\label{tab:tab6}
\resizebox{.8\textwidth}{!}{%
\begin{tabular}{@{}llllll@{}}
\toprule
\textbf{Author} & \textbf{Techniques} & \textbf{Accuracy} & \textbf{Precision} & \textbf{Recall} & \textbf{F1-score} \\ \midrule
\textbf{~\cite{sharma2022smotified}} & SMOTified-GAN & 90.75\% & 1 & 1 & 1 \\
\textbf{~\cite{siddappa2019adaptive}} & LMDL & 80.95\% & .80 & .81 & .79 \\
\textbf{~\cite{mohamad2021improving}} & PCA-Ranker & 77.68\% & 0.44 & 0.37 & 0.38 \\
\textbf{Our study} & BSGAN & 99.29\% & 0.9786 & 0.9785 & 0.9783 \\ \bottomrule
\end{tabular}%
}
\end{table}


On the Abalone dataset, as shown in Table~\ref{tab:tab7}, BSGAN becomes the second-best algorithm in terms of precision, recall, and F1-score, while the question raised as SMOTified-GAN demonstrates nearly perfect precision, recall, and F1-score.

\begin{table}[h!]
\centering
\caption{Comparison with the previous study on Abalone datasets.}
\label{tab:tab7}
\resizebox{.8\textwidth}{!}{%
\begin{tabular}{@{}llllll@{}}
\toprule
\textbf{Author} & \textbf{Techniques} & \textbf{Accuracy} & \textbf{Precision} & \textbf{Recall} & \textbf{F1-score} \\ \midrule
\textbf{~\cite{sharma2022smotified}} & SMOTified-GAN & 98.61\% & 1 & 1 & 0.9222 \\
\textbf{~\cite{jo2022obgan}} & - & - & - & 0.5960 & 0.4908 \\
\textbf{~\cite{mohamad2021improving}} & PCA-Ranker & 99.23\% & 0.5 & 0.5 & 0.5 \\
\textbf{~\cite{jadhav2020novel}} & svmradial & 97.70\% & 0.00 & - & 0.00 \\
\textbf{Our study} & BSGAN & 94.18\% & 0.9049 & 0.9064 & 0.9052 \\ \bottomrule
\end{tabular}%
}
\end{table}

On Wine quality datasets as shown in Table~\ref{tab:tab78}, BSGAN became the second-best algorithm in terms of precision, recall, and F1-score, while PCA-Ranker showed the best results. Again with 97.19\% accuracy achieving a nearly perfect score of precision, recall, and F1-score is hardly feasible.

\begin{table}[h!]
\centering
\caption{Comparison with the previous study on Wine quality datasets.}
\label{tab:tab78}
\resizebox{.8\textwidth}{!}{%
\begin{tabular}{llllll}
\hline
\textbf{Author} & \textbf{Techniques} & \textbf{Accuracy} & \textbf{Precision} & \textbf{Recall} & \textbf{F-1 score} \\ \hline
\textbf{~\cite{jo2022obgan}} & OBGAN & - & - & 0.5389 & 0.6508 \\
\textbf{~\cite{siddappa2019adaptive}} & LMDL & 71.11\% & .72 & .71 & .71 \\
\textbf{~\cite{mohamad2021improving}} & PCA-Ranker & 97.19\% & 1 & 1 & 1 \\
\textbf{~\cite{sharma2022smotified}} & SMOTified-GAN & 95.58\% & 0.53 & 0.69 & 0.5274 \\
\textbf{Our study} & BSGAN & 93.90\% & 0.9390 & 1.0 & 0.9685 \\ \hline
\end{tabular}%
}
\end{table}


    


\begin{figure}[h!]
    \centering
    \includegraphics[width=.8\textwidth]{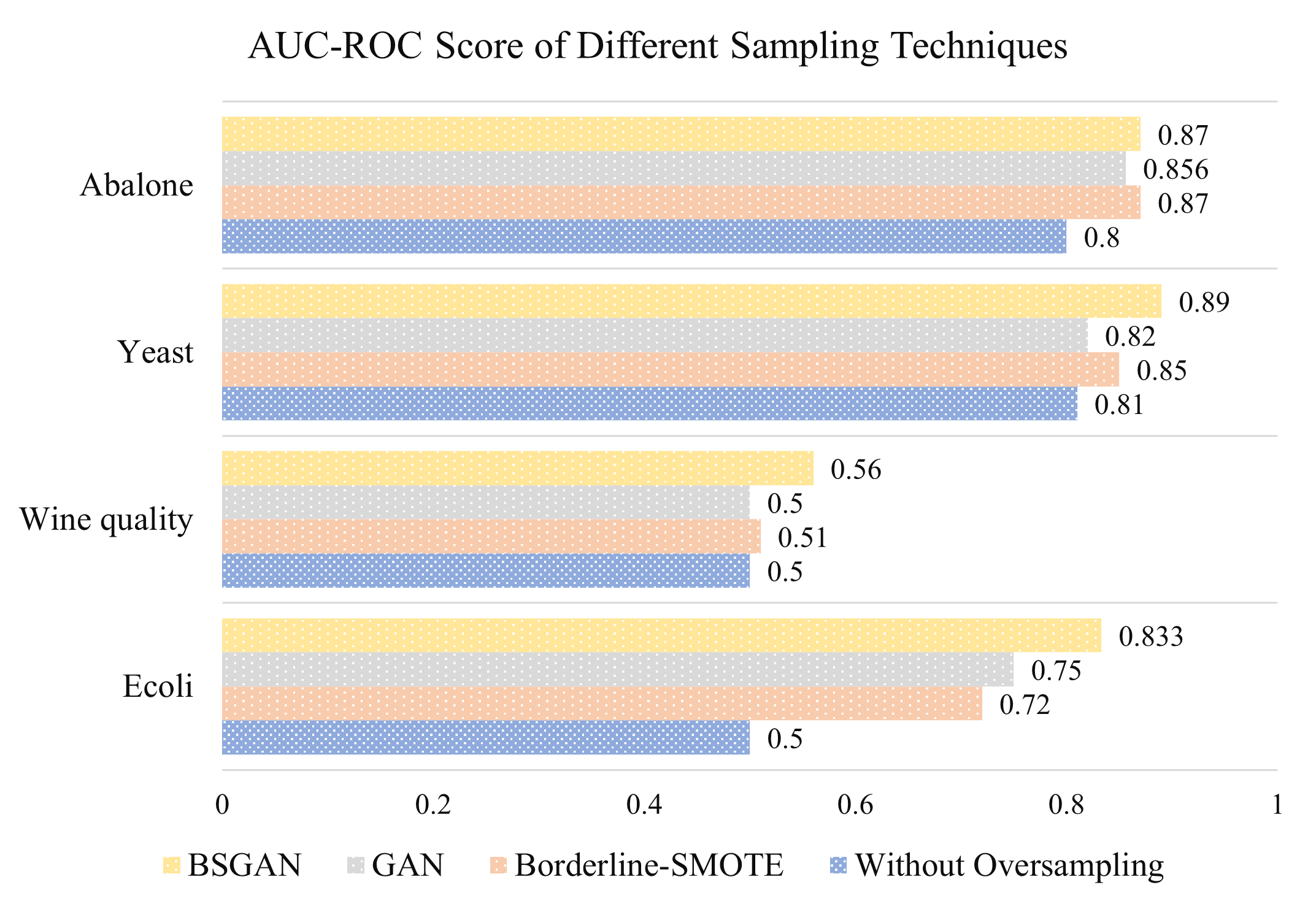}
    \caption{AUC-ROC scores for different sampling techniques on referenced imbalanced datasets used in this study.}
    \label{fig:areac}
\end{figure}

In Figure~\ref{fig:areac}, measures of the Area Under the Curve (AUC) of the Receiver Characteristics Operator (ROC) are plotted for each oversampling technique applied to the test set of different datasets. Our proposed BSGAN shows the best performance on all datasets, and the highest AUC score (0.89) is achieved on the yeast dataset. The worst performance (AUC = 0.5) is achieved on Wine quality and Ecoli datasets without applying any oversampling techniques.

\begin{figure}[h!]
    \centering
    \includegraphics[width=.9\textwidth]{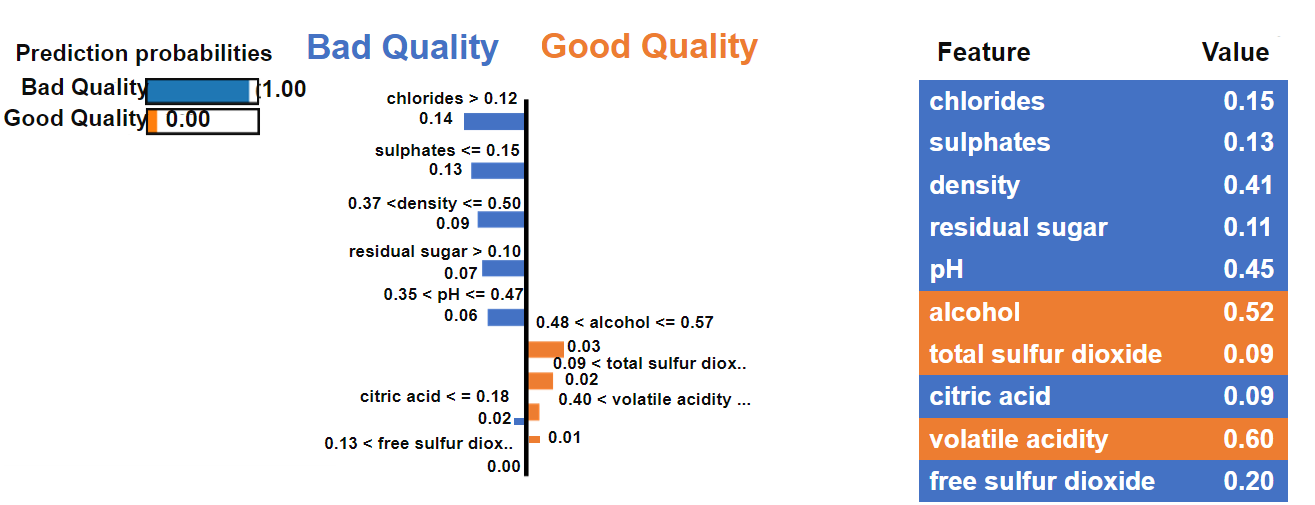}
    \caption{Interpreting the model using LIME on the Wine quality dataset.}
    \label{fig:lime}
\end{figure}

During the study, Local Interpretable Model-Agnostic Explanations (LIME) were employed to assess the black box behavior of our proposed models. LIME, a valuable tool for model interpretability, affords us an understanding of the rationales behind the predictions made by the model through analysis and visualization of the individual feature contributions. This is illustrated in Figure~\ref{fig:lime}, which shows various features' contributions to the Wine quality prediction. The model is 99\% confident that the predicted Wine Quality is poor, and the variables with the most significant impact on the predicted wine quality are Sulfate, Sulfur dioxide, volatile acidity, and chloride.

\begin{figure}[h!]
    \centering
    \includegraphics[width=.9\textwidth]{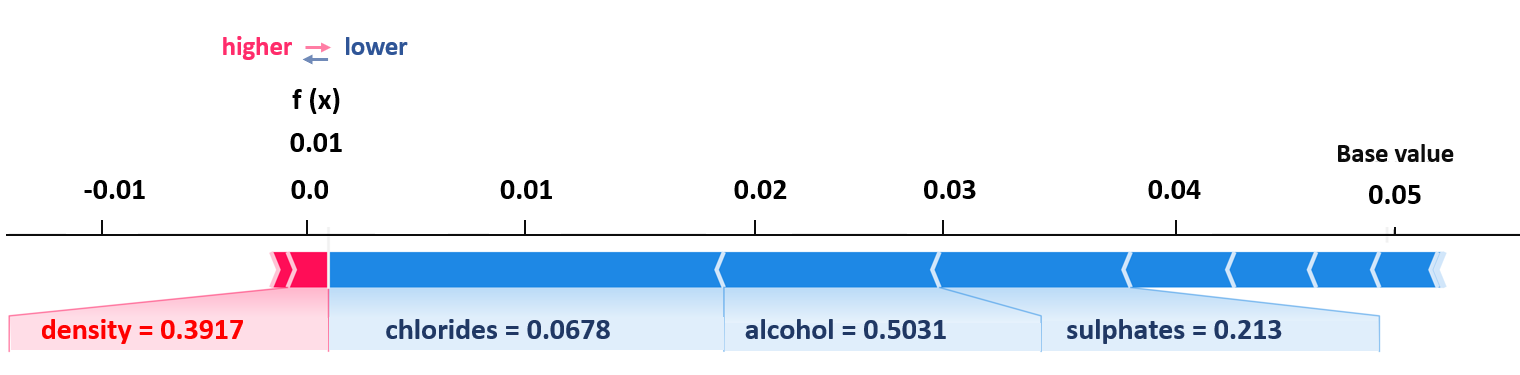}
    \caption{Force Plot observation of the Wine quality data using SHAP.}
    \label{fig:force}
\end{figure}

Additionally, the Shapley Additive Explanations (SHAP) framework was employed to comprehend the prediction outcomes of the model on the oversampled dataset with more clarity. As depicted in Figure~\ref{fig:force}, the illustration presents a forced plot of the first observation in the Wine quality dataset. This force plot graphically illustrates the influence of each feature on the prediction made by the model. The figure shows that the baseline value is 0.3, and the final value, f(x) = 0.76, signifies the predicted value of the abalone.

\begin{figure}[h!]
    \centering
    \includegraphics[width=.8\textwidth]{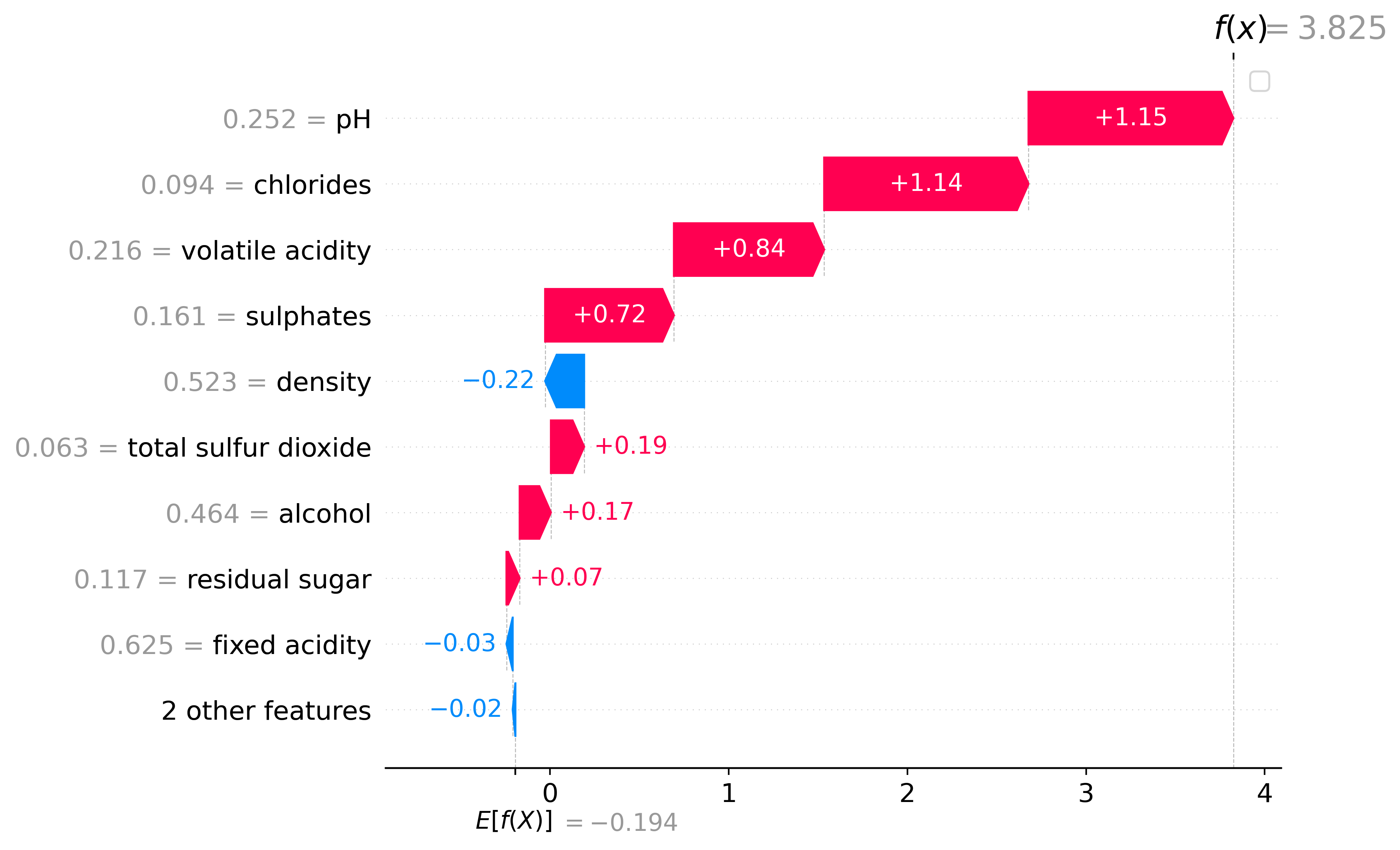}
    \caption{A Waterfall plot example for the median predicted wine quality in the Wine quality dataset.}
    \label{fig:waterfall}
\end{figure}

Figure~\ref{fig:waterfall} presents a SHAP explanation for the second observation in the test data from the Wine quality dataset. The actual outcome reflects poor wine quality, which the model accurately predicted. The figure displays the average predicted score of the dataset, represented by E(f(x)), at the bottom and is equal to -0.194. The prediction score for the specific instance, represented by f(x), is shown at the top and equals 3.825. The waterfall plot sheds light on the contribution of each feature in the prediction process, leading to a change in the prediction from E(f(x)) to f(x). The feature `pH' is seen to have the most significant impact and plays a crucial role in the prediction by decreasing the prediction value. Conversely, the feature `density' has a negative impact on the prediction outcome.

\begin{figure}[h!]
    \centering
    \includegraphics[width=.8\textwidth]{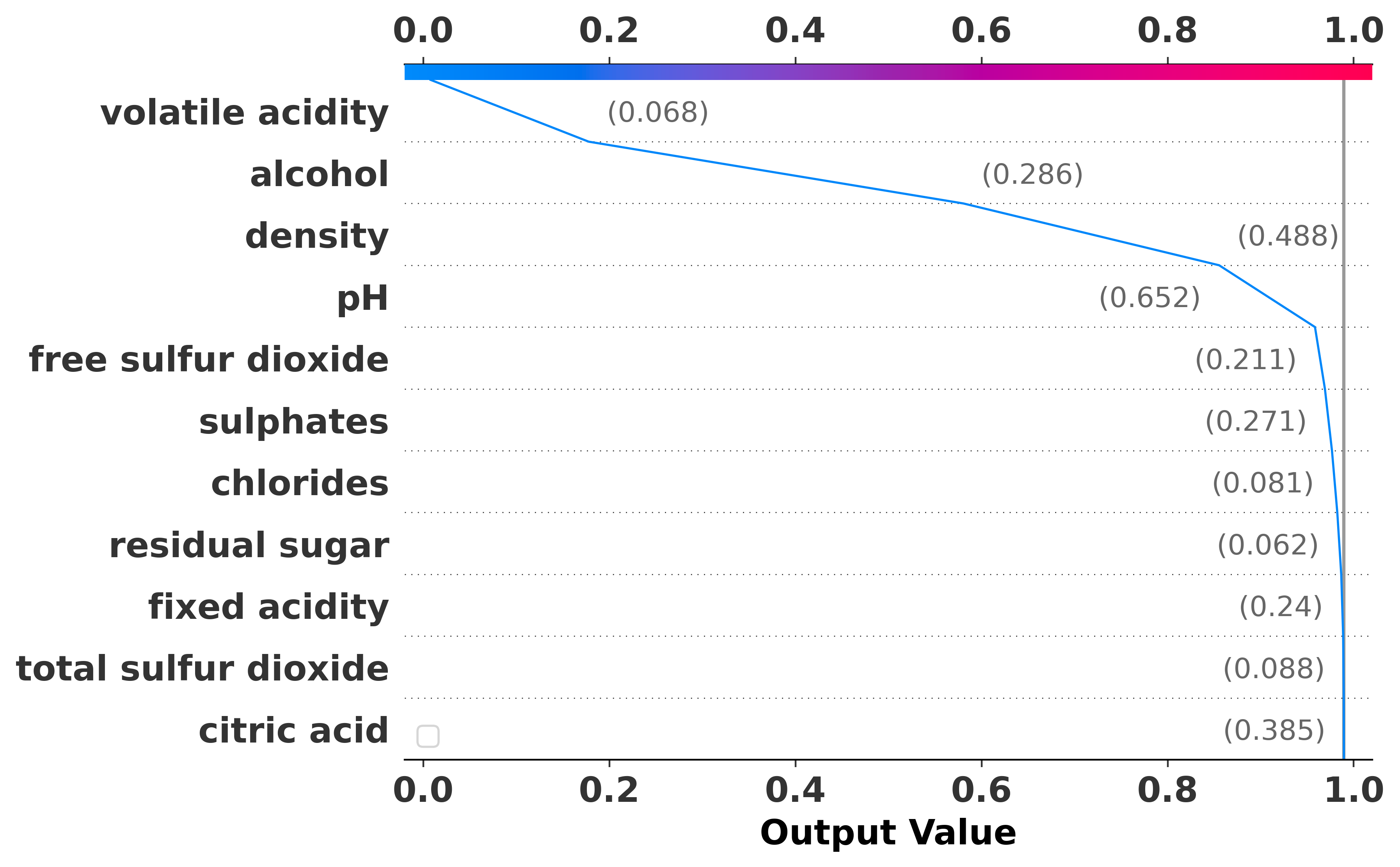}
    \caption{Model interpretation with expected value using SHAP on Wine quality dataset.}
    \label{fig:expectedfinal}
\end{figure}

Figure~\ref{fig:expectedfinal} presents a SHAP explanation of the 15th observation in the test data from the Wine Quality dataset. The actual outcome depicts a good-quality wine, which the model correctly predicted. As seen in the figure, the expected value is near 1, indicating that factors such as pH and citric acid played a significant role in the model's determination of the wine as being of good quality.

\section{Conclusions}\label{con}
Our study proposed and assessed the performance of BSGAN approaches to handle the class imbalanced problems using four highly imbalanced datasets. We revealed that our proposed approach outperformed Borderline-SMOTE and GAN-based oversampling techniques in various statistical measures. Additionally, the comparison between our state-of-the-art techniques using neural network approaches outperformed many of the existing proposed recent reference approaches, as highlighted in Tables~\ref{tab:tab5}--~\ref{tab:tab78}. The inter-class distance measurement ensures that the data distribution follows Gaussian distribution after data expansion using BSGAN, as referred to in Table~\ref{tab:tab4}. The findings of the proposed techniques should provide some insights to researchers and practitioners regarding the advantage of GAN-based approaches and help to understand how they can potentially minimize the marginalization and sensitivity issues of the existing oversampling techniques. Future works include but are not limited to applying BSGAN on other high imbalance and big datasets, experimenting with mixed data (numerical, categorical, and image data), changing the parameters of the proposed models, and testing it for multiclass classification.

\bibliographystyle{unsrt}  
\bibliography{main}

\end{document}